\documentclass{article}

\usepackage[final,nonatbib]{crl_neurips_2023}
\usepackage[numbers]{natbib}

\usepackage[utf8]{inputenc} 
\usepackage[T1]{fontenc}    
\usepackage{hyperref}       
\usepackage{url}            
\usepackage{booktabs}       
\usepackage{amsfonts}       
\usepackage{nicefrac}       
\usepackage{microtype}      
\usepackage{xcolor}         
\usepackage{amsmath}
\usepackage{amssymb}
\usepackage{graphicx}
\usepackage{diagbox}
\usepackage[flushleft]{threeparttable}
\usepackage{color}

\title{Attention for Causal Relationship Discovery from Biological Neural Dynamics}

\author{%
  Ziyu Lu\thanks{This work was done when Ziyu Lu was a summer intern at Oak Ridge National Laboratory under the NSF-MSGI program.} \\
  University of Washington\\
  Seattle, WA 98195 \\
  \texttt{luziyu@uw.edu} \\
  \And
  Anika Tabassum \\
  Oak Ridge National Laboratory\\
  Oak Ridge, TN 37830 \\
  \texttt{tabassuma@ornl.gov} \\
    \And
  Shruti Kulkarni  \\
  Oak Ridge National Laboratory\\
  Oak Ridge, TN 37830 \\
  \texttt{kulkarnisr@ornl.gov} \\
  \And
  Lu Mi \\
  Allen Institute for Brain Science\\
  Seattle, WA 98109 \\
University of Washington\\
  Seattle, WA 98195 \\
  \texttt{lu.mi@alleninstitute.org} \\
  \And
  J. Nathan Kutz  \\
  University of Washington\\
  Seattle, WA 98195 \\
  \texttt{kutz@uw.edu} \\
  \And
  Eric Shea-Brown \\
  University of Washington\\
  Seattle, WA 98195 \\
  \texttt{etsb@uw.edu} \\
    \And
  Seung-Hwan Lim  \\
  Oak Ridge National Laboratory\\
  Oak Ridge, TN 37830 \\
  \texttt{lims1@ornl.gov} \\
}

\begin{document}

\maketitle

\begin{abstract}
  This paper explores the potential of the transformer models for learning Granger causality in networks with complex nonlinear dynamics at every node, as in neurobiological and biophysical networks. Our study primarily focuses on a proof-of-concept investigation based on simulated neural dynamics, for which the ground-truth causality is known through the underlying connectivity matrix. For transformer models trained to forecast neuronal population dynamics, we show that the cross attention module effectively captures the causal relationship among neurons, with an accuracy equal or superior to that for the most popular Granger causality analysis method. While we acknowledge that real-world neurobiology data will bring further challenges, including dynamic connectivity and unobserved variability, this research offers an encouraging preliminary glimpse into the utility of the transformer model for causal representation learning in neuroscience.  
\end{abstract}

\section{Introduction}
Our abilities to learn, think, and create are all shaped by the nonlinear dynamics of massive numbers of neurons, orchestrated through their connections. Obtaining a comprehensive map of this connectivity, or in other words, a connectome, is a major endeavor in neuroscience \cite{sporns2011human}, with applications including neurological and psychiatric disorders \cite{van2019cross} and developing brain-machine interfaces \cite{monti2015measuring}. Building a connectome is a very challenging task, for starters because of the vast number of neurons and possible connections in any brain or brain system. Nevertheless, stunning recent advances in, e.g., automated 2-D to 3-D image reconstruction from electron microscopy (EM) \cite{EM}, have led to breakthroughs in mapping the structure of synaptic connectomes~\cite{microns2021functional, scheffer2020connectome, mahalingam2022scalable}. This said, there are substantial limitations in detecting long-range connections in mammalian-scale brains due to tissue segmentation. Moreover, EM and allied imaging techniques deliver a static view of the structure of connectivity after the sacrifice of the animal. However, in living networks, connectivity is dynamic, due to multiple plasticity, neuromodulation, and physiological mechanisms \cite{bargmann2013connectome}.

An alternative to structural approaches is to infer connectivity from recordings of neural activity \cite{perkel1967neuronal, ostojic2009connectivity, siegle2021survey}. Recent advances in electrophysiological recording technologies \cite{steinmetz2021neuropixels} have enabled simultaneous and long-term monitoring of large neuronal populations across multiple brain regions, offering a richer functional characterization than ever. This presents a new opportunity to develop computational methods capable of extracting neuronal network structure from functional data.

In fact, the need to uncover network structure from observed network dynamics resonates across various fields beyond neuroscience, such as traffic prediction and weather forecasting. Building upon the concept of Granger causality \cite{granger1969investigating}, myriad methods have been deployed toward this challenge, from classical statistical reasoning to recent deep learning techniques. However, it is not clear that these approaches are well matched to the complexities of neurobiological networks. Once again, the non-stationarity of neuronal responses \cite{tomko1974neuronal}, especially when driven by external stimuli that dramatically alter neurons' firing rates and their sensitivity to inputs, poses a significant obstacle. Meanwhile, neuronal connectivity is not static and can evolve over multiple timescales \cite{abbott2000synaptic}, adding another layer of complexity. Furthermore, the multiscale connectivity of the brain, across both time and space, ensures the presence of unobserved sources of variability, as all recordings represent only a fraction of the brain's neurons and will miss many determinants of its evolution over time. Overall, neuroscience datasets are often of a substantial scale and are ever-growing, necessitating models with robust scalability. And lastly, observations of neuronal networks can be sparse, particularly in the case of binary spike trains where information is communicated via occasional spikes, demanding models capable of handling long-term history dependencies. These multifaceted challenges collectively underscore the need for tailored approaches to unraveling connectivity in neurobiological networks. 

Recently, the transformer model has demonstrated remarkable efficacy in capturing intricate relationships in sequence-to-sequence modeling \cite{brown2020language}, surpassing recurrent neural networks in managing long-term dependencies while preserving the structural interpretability featured in graph neural networks. In this study, we explore the potential of the transformer models for
causal representation learning in neuroscience. Our contributions are as follows. (1) We propose a novel approach, Causalformer, for discovering Granger causal relationships among neurons using the cross attention matrix. (2) With Causalformer, we highlight the key factors needed to enable a clear causal interpretation from forecasting models. (3) We validate our approach on synthetic neuronal datasets, and achieve performance equal or superior to that for the most popular Granger causality analysis method. 

\section{Related Work}
\textbf{Granger causality}. In neuroscience, Granger causality has been the most widely used framework for identifying directed functional interactions between neurons \cite{seth2015granger}. Given activities (such as membrane potential or spike trains) of two neurons $i$ and $j$, if predictions of neuron $j$'s future can be improved when given neuron $i$'s history besides neuron $j$'s own history, then neuron $i$ is said to Granger-cause neuron $j$. The classical, and still most popular, method for identifying Granger causal relationships uses the vector autoregression model (VAR). The VAR approach has been favored for its simplicity and scalability, and standard tools such as the Multivariate Granger Causality (MVGC) Matlab Toolbox~\cite{barnett2014mvgc} have been developed. Nevertheless, the VAR approach can struggle when the time series is non-stationary or when the temporal relationship cannot be well captured in a linear fashion, as in the case of neuronal interactions. 

\textbf{RNNs}. To account for more complex and nonlinear dynamics, recently deep learning models have been employed for identifying causal relationships. \cite{tank2021neural} proposed approaches using multilayer perceptrons (MLPs) and recurrent neural networks (RNNs). In order to avoid the confusion of shared hidden units when all objects are modeled simultaneously, they modeled the dynamics of each object with a separate MLP or RNN. An immediate concern for this approach is that it may struggle to scale to large neuronal populations due to its need to optimize a separate model for each neuron. In addition, it may be hard to train RNNs to capture long-term history dependencies, due to issues such as catastrophic forgetting and gradient exploding or vanishing.

\textbf{GNNs}. To allow for the simultaneous modeling of the dynamics of all objects in the system while maintaining a disentangled representation of the information flow, 
\cite{lowe2022amortized} proposed an approach based on graph neural networks (GNNs). They adopted an encoder-decoder architecture, with the encoder learning a causal graph that is static over time, and the decoder learning the dynamics that drive the temporal evolution of the system. Since neuronal interactions can be dynamic and time-dependent, the separation of causal graph learning from dynamics learning may create an information bottleneck. 

Due to the aforementioned concerns, the previous deep learning approaches may not be particularly suitable for modeling neuronal causal relationships. In fact, to the best of our knowledge, none of them has been applied to biological neural dynamics. Nonetheless, they demonstrate the exciting possibility of leveraging deep learning tools for causal discovery, which motivates our usage of the transformer model. Besides being able to perform multivariate time series forecasting with high accuracy \cite{zhou2021informer, grigsby2021long, zhang2022crossformer}, the transformer model has been shown to capture long-range history dependencies better than RNNs \cite{bahdanau2014neural, vaswani2017attention, zaheer2020big}, while allowing for similar interpretability as GNNs~\cite{battaglia2018relational}. Meanwhile, techniques for handling non-stationary time series \cite{liu2022non} and further improving the scalability of transformer models \cite{choromanski2020rethinking} have been developed. Therefore, transformer models stand out to us as a promising candidate for identifying neuronal causal relationships. In fact, the transformer model has been successfully applied to model neuronal dynamics by \cite{le2022stndt} in a non-causal setting. 

\section{Model and Results}
\begin{figure}
  \centering
  \includegraphics[width=1.0\textwidth]{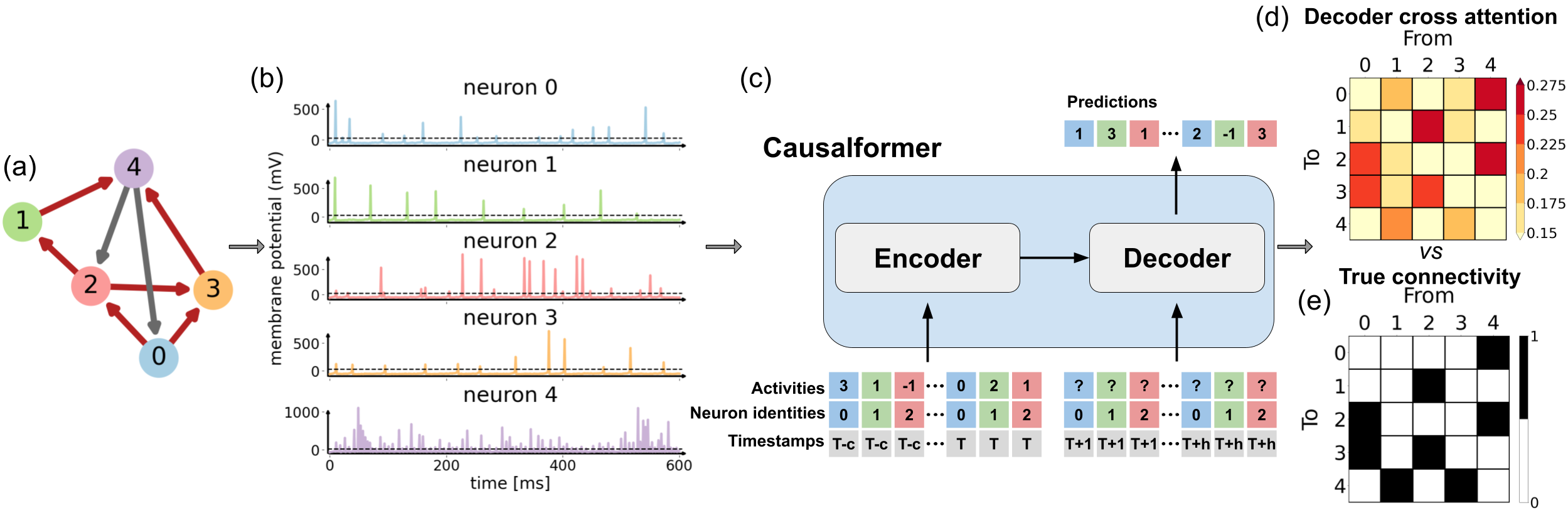}
  \caption{Experiment overview. \textbf{(a)} Example five-neuron network connectivity structure. Red arrows represent excitatory connections, and gray arrows represent inhibitory connections. \textbf{(b)} Example membrane potential of the neuronal population in (a), simulated with the Izhikevich model. Firing threshold is marked by dashed black horizontal lines. \textbf{(c)} Illustration of the encoder-decoder architecture of Causalformer. \textbf{(d)} Example decoder cross attention matrix from Causalformer trained on the membrane potential of (a). \textbf{(e)} Ground-truth connectivity matrix for (a). Entry $(i, j)$ is 1 if there is a connection from neuron $j$ to $i$ (regardless of inhibitory or excitatory), and 0 otherwise.}
  \label{diagram}
\end{figure}

We demonstrate the transformer model's capability of discovering underlying causal relationship between neurons on simulated neuronal datasets, where the ground-truth connectivity patterns are known. As illustrated in Figure \ref{diagram}, on each dataset, we trained a transformer model to predict the future activities based on the past, and examined the degree to which the trained transformer model -- specifically, its cross attention matrix -- corresponds to the ground-truth connectivity. 

This section is organized as follows. In section 3.1, we describe our transformer model. In section 3.2, we briefly review the generation of synthetic datasets. In section 3.3, we present the results. 

Our code is available at \url{https://github.com/ziyulu-uw/causalformer}. 

\subsection{The Causalformer}
Following \cite{vaswani2017attention}, numerous transformer architectures have been proposed for different purposes, including multivariate time series forecasting \cite{zhou2021informer, lim2021temporal, grigsby2021long}. As Granger causality between two objects $i$ and $j$ is based on whether the time series of object $i$ can help predicting the time series of object $j$, we propose to perform causal discovery by extending transformers for multivariate time series forecasting. Specifically, based on the Spacetimeformer proposed by \cite{grigsby2021long}, we propose the Causalformer, which can jointly forecast the future activities of multiple objects and learn the causal relationship among them (see Figure \ref{diagram}(c) for a sketch and Appendix Figure~\ref{model} for details). 

Causalformer preserves the signature spatiotemporal attention mechanism of the Spacetimeformer and follows the same encoder-decoder architecture. In the case of neural dynamics, we describe the Causalformer as follows.  Let the activities of $N$ neurons form an $N$-dimensional time series $\{x_t\}_{t=0, 1, 2, \dots, T}$, where $x_t = (x_{t1}, \dots, x_{tN})^\top \in \mathbb{R}^N$ is the activities of all neurons at time $t$, and $x_{ti}$ is the activity of neuron $i$ at time $t$. Given a $c+1$-step $N$-dimensional history sequence $(x_{t-c}, \dots, x_t)$, Causalformer is tasked to predict an $h$-step (future) target sequence $(x_{t+1}, \dots, x_{t+h})$. In the spatiotemporal attention mechanism, the history sequence $(x_{t-c}, \dots, x_t)$ is flattened from an $(N, c+1)$ array to an $(N(c+1), 1)$ array, where each timestep of each neuron is embedded as a separate token. The encoder takes in the embedding of the history timestamps $(t-c, \dots, t)$, the neuron identities $(1, \dots, N)$, as well as the flattened history sequence, and learns a representation for each history timestep of each neuron through a ``local'' self attention module, where each neuron is only allowed to attend to its own history. The representations are further processed through a point-wise feed-forward network, which ``is applied to each position separately and identically” \cite{vaswani2017attention}. Importantly, since neurons are not allowed to attend to other neurons' history in the self attention module, in the history representations learned by the encoder, neuron $i$'s representations only contain information about its own past. 

On top of the encoded representations of history neural activities, the decoder learns relationships across neurons across time, which can be interpreted as the learned Granger causality. The decoder takes in the embedding of the target timestamps $(t+1, \dots, t+h)$, neuron identities $(1, \dots, N)$, and the zero-filled flattened target sequence, together with the history representations learned by the encoder. The representation of the target sequence is first updated through a cross attention module, and then through a point-wise feed-forward network. In the cross attention module, the queries are linear projections of the target sequence embedding, and keys and values are linear projections of the encoder-learned history representations (see Appendix~\ref{sec:attn} for more details). Therefore, all information, if there is any, that is passed from neuron $i$'s history to neuron $j$'s future is captured in the decoder cross attention, which effectively captures Granger causality. 

Graphically, consider a directed graph consisting of $2N$ nodes, including $N$ history and $N$ future nodes representing the history and future of each neuron, respectively, and edges going from history nodes to future nodes. Following \cite{singh2023attention}, the $N \times N$ decoder cross attention matrix can be seen as a posterior distribution over the possible edges in the graph given the history and the future nodes, with the $j$-th row of the matrix representing the posterior distribution over the $N$ possible edges between neuron $j$'s future and all neurons' history. The existence of an edge going from neuron $i$'s history node to neuron $j$'s future node then corresponds to a Granger causal relationship between neuron $i$ and $j$. Here, due to flattening, the decoder cross attention matrix actually has dimensions $Nh \times N(c+1)$, but we can aggregate across the timesteps for each neuron to determine the causal relationship. For instance, if there exists an edge from any history step of neuron $i$ to any future step of neuron $j$, then we say that neuron $i$ Granger-causes neuron $j$.

\subsection{Data generation}
Since the inter-neuron connection pattern is hardly available in in vivo neuronal recordings, we applied the Causalformer to synthetic data where we can access the ground-truth connectivity. To make the synthetic data exhibit dynamics similar to that of real neurons, we simulated the data following the Izhikevich model \cite{izhikevich2003simple}, which has been widely employed to generate data for evaluating Granger causality methods\cite{cadotte2008causal}. Briefly, the model produces nonlinear dynamics from a two-dimensional system of differential equations, 
\begin{align*}
    v' &= 0.04v^2 + 4.1v + 108 - u + I \\
    u' &= a(bv-u)
\end{align*}
with $30mV$ as the spiking threshold and an auxiliary after-spike resetting mechanism
\begin{equation*}
    \text{if } v \geq 30mV, \text{then } \begin{cases}
        v \leftarrow c \\
        u \leftarrow u + d
    \end{cases}
\end{equation*}
The variable $v$ represents the membrane potential of the neuron, and the variable $u$ is a membrane recovery variable, accounting for $K^+$ activation and $Na^+$ inactivation in the neuron. When a neuron spikes, it transmits electrical signals through synapses to other neurons it connects to. The synaptic currents, together with externally injected currents, are captured in variable $I$. The part $0.04v^2 + 4.1v + 108$ is suggested by \cite{izhikevich2003simple}, and was obtained by fitting the dynamics of real cortical neurons. At the beginning of the simulation, $v$ is initialized as $v_0 = -65mV$, and $u$ is initialized as $u_0 = bv_0$. Each time-step of the simulation corresponds to 1ms. 

\begin{figure}
  \centering
  \includegraphics[width=1.0\textwidth]{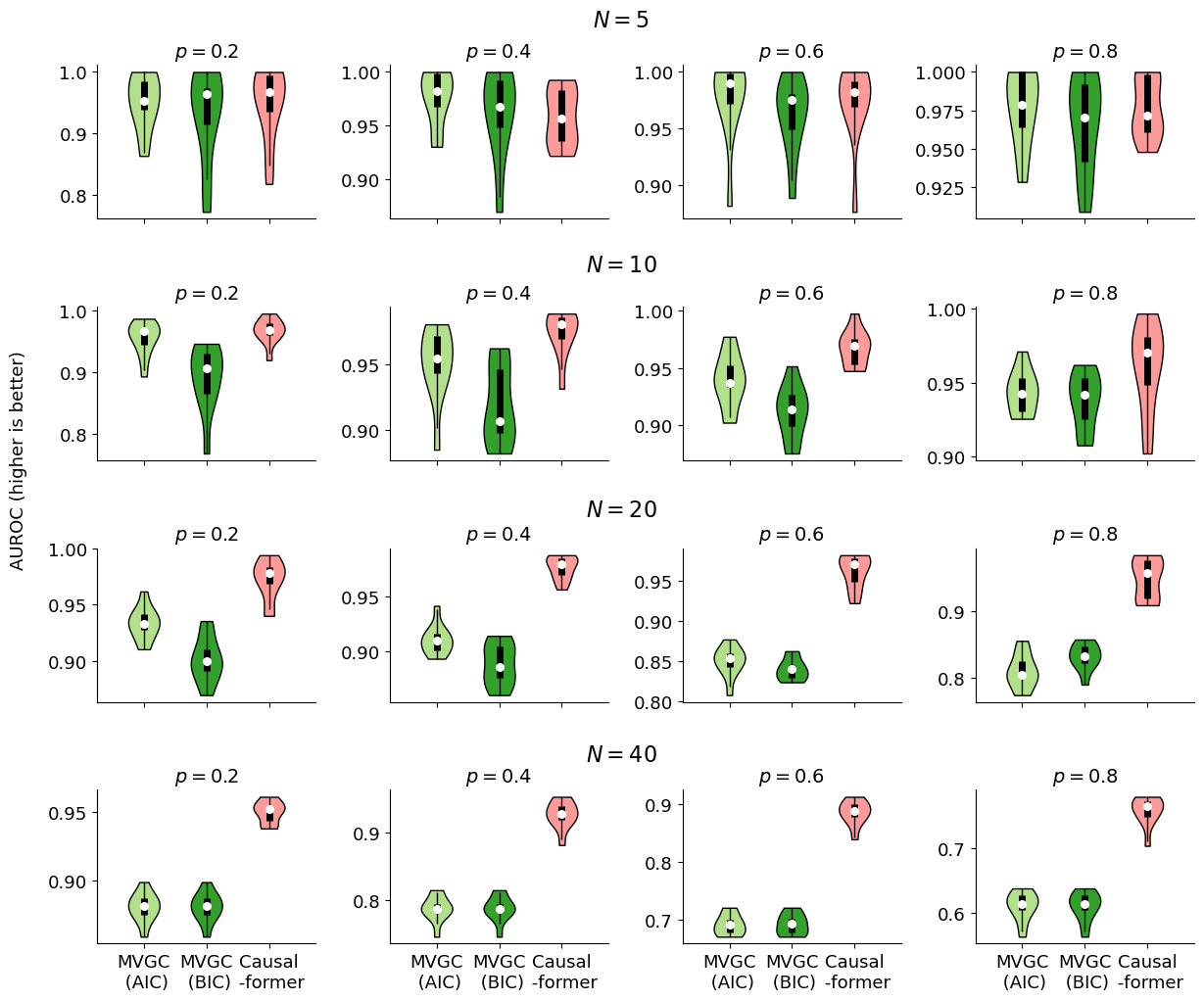}
  \caption{Results with random networks of 4 different sizes $N=5, 10, 20, 40$ and 4 different connectivity probabilities $p=0.2, 0.4, 0.6, 0.8$. 10 different typologies were simulated for each $N$ and $p$. Violin plots show the empirical distributions estimated from the 10 random networks. White dots mark the medians of the distributions, and black dots mark the 25th to the 75th percentile. Y-axis shows AUROC (between 0 and 1, higher is better). X-axis shows the methods used for inferring causality. \texttt{MVGC (AIC)}, \texttt{MVGC (BIC)}, \texttt{Causalformer} stand for Multivariate Granger Causality with VAR model order selected by AIC and BIC, and Causalformer decoder cross attention, respectively.}
  \label{randnets}
\end{figure}

\subsection{Results}
We simulated random networks of size $N = 5, 10, 20, \text{and } 40$, each with both excitatory and inhibitory neurons, at a ratio of 4:1, following \cite{izhikevich2003simple} (see Figure \ref{diagram}(a) for an example network). Additional parameters are listed in Appendix Table~\ref{izh}. To create baseline dynamics, we injected a noisy input $I_{\text{inp}} \sim N(\mu=5, \sigma^2=5)$ to each neuron at every timestep. The connectivity strength between neurons was set to 5. That is, when there is an excitatory connection from neuron $i$ to $j$, if neuron $i$ spikes at time $t$, the input to neuron $j$ at time $t+1$ will be $I = 5 + I_{\text{inp}}$, or $I = -5 + I_{\text{inp}}$ if the connection is inhibitory. All edges are directed, and are connected randomly and independently with probability $p$. Edges from neuron $i$ to itself are ignored, since the dependence on its own history is already coded in the dynamics. For each network size, we experimented with $p = 0.2, 0.4, 0.6, \text{and } 0.8$. 10 different networks were simulated for each network size and each connectivity probability $p$, resulting in a total of 160 different random networks. 

For each random network, we ran the simulation for 5000 time steps, and used the first 60\% of steps for training, the next 20\% for validation, and the last 20\% for testing. Models were trained on the normalized membrane potential (see Figure \ref{diagram}(b) for example traces), which was first clipped at the spiking threshold $(30mV)$ and then standardized with a z-score transformation. For each random network, we trained Causalformer with 10 different random seeds to predict $x_{t+1}$ (i.e., $h=1$) from $x_{t}, \dots, x_{t-c-1}$, where $x_{t} \in \mathbb{R}^N$ denotes the normalized membrane potential of all neurons at time $t$, and we set the history horizon $c=10$. We used the same set of hyperparameters across all Causalformer models, which can be found in Appendix Table~\ref{hyper}. 

After training, the models were evaluated on the test set, and the analysis on decoder cross attention matrices was also performed on test set samples. As a sanity check, we first confirmed qualitatively (see Appendix Figure \ref{pred}) that the models were able to capture the dynamics reasonably well, for otherwise the cross attention may not be very trustworthy. Each test sample has a set of decoder cross attention matrices $A^{(k)} \in \mathbb{R}^{N \times Nc}, k=1, \dots, K$, where $K$ is the number of attention heads in the model. For each attention head, we aggregated across the history steps of each neuron to obtain $\hat{A}^{(k)} \in \mathbb{R}^{N \times N}$, where $\hat{A}^{(k)}_{ji} = \text{sum}(A^{(k)}[j, ic:(i+1)c])$, representing the probability of the existence of an edge from any history step of neuron $i$ to the future step of neuron $j$. Then we aggregated across all attention heads, obtaining $\hat{A} = \sum_{k=1}^K \hat{A}^{(k)}_{ji}$. We computed $\hat{A}_{\text{sampavg}}$ as the average $\hat{A}$ across all test samples, and renormalized such that every row sum up to 1. Lastly, we took $\hat{A}_{\text{modelavg}}$ as the average $\hat{A}_{\text{sampavg}}$ across the 10 models initialized with different random seeds. 

In previous works \cite{tank2021neural, lowe2022amortized}, whether or not object $i$ Granger-causes object $j$ is decided by whether the history of object $i$ is weighed by zero in the prediction of object $j$. However, because of the softmax operation in attention computation, the decoder cross attention matrix will not contain exact zeros. Therefore, instead of zero, we seek positive scalars as the threshold. That is, when the probability of the existence of an edge from neuron $i$ to $j$ is greater than the threshold, we say that neuron $i$ Granger-causes neuron $j$. To avoid the arbitrariness in the threshold selection, when evaluating the correspondence between the inferred and true connectivity, we used the Area Under the Receiver Operating Characteristic (AUROC) as used in \cite{lowe2022amortized}, bypassing the selection of a specific threshold (see details in Appendix~\ref{sec:auroc}). AUROC for Causalformer was computed by comparing $\hat{A}_{\text{modelavg}}$ to the ground-truth connectivity matrix. Note that since there are no self-edges, the diagonal entries of the ground-truth connectivity matrix are zero. Therefore, when computing AUROC, the diagonal entries of $\hat{A}_{\text{modelavg}}$ were also replaced with zeros. 

For comparison, we also estimated the pairwise-conditional Granger causality using the VAR model implemented in the MVGC toolbox \cite{barnett2014mvgc}, which also takes the form of an $N \times N$ matrix, where the $(j, i)$-th entry is the likelihood that neuron $i$ Granger-causes neuron $j$. AUROC scores were computed between the likelihoods reported by MVGC (with diagonal entries set to zero) and the ground-truth connectivity. For VAR model order selection, we used either AIC or BIC~\cite{mcquarrie1998regression}, and both results were reported. Figure \ref{randnets} compares AUROC scores achieved by MVGC and Causalformer on the random networks. The decoder cross attention of Causalformer is able to reflect the underlying connectivity to a similar degree as the MVGC method for small $N$ ($N = 5, 10$), and has a clear advantage over MVGC for larger $N$ ($N = 20, 40$). We speculate that increasingly complicated connectivity patterns in larger networks can amplify the nonlinear aspect of neuronal population dynamics, which can be captured by Causalformer through its nonlinear feed-forward transformations, but may not be well captured by the linear VAR model in MVGC. This exciting trend aligns with our ultimate goal of discovering causal relationships among neurons from large-scale neuronal population recordings.

\section{Discussion}
We demonstrate the potential of discovering Granger causal relationships from transformer models trained for multivariate time series forecasting in a neuroscience setting. With Causalformer, we stress that in order for the decoder cross attention matrix to reflect Granger causality, it is key to ensure that history information of neurons is not mixed in the encoder and future information is not mixed in the decoder. Beyond the Causalformer, these ideas can be easily extended to other transformers with a similar encoder-decoder architecture. In fact, incorporating this idea to transformers which are able to better handle non-stationarity~\cite{liu2022non} or with better scalability~\cite{choromanski2020rethinking} is an important direction for our future work. With the ability to record more neurons simultaneously than ever, the causal perspective on the interpretation of cross attention is of particular importance to the neuroscience community, as transformer models present a potential solution to model the nonlinear dynamics of large neuronal populations and their long, complex history dependencies. 

While this work provides a proof of concept on synthetic datasets, several limitations must be addressed in the future work. To start with, in this work the neural dynamics were characterized by continuous neuronal membrane potential, but the model's robustness to sparse inputs, such as spike trains, remains to be tested. Furthermore, although we have briefly touched on the advantage of transformer methods over previous VAR, MLP, RNN, or GNN based approaches, a thorough comparison with more networks and more metrics, such as those used in \cite{lusch2016inferring}, is much needed to understand the limitations of each model. Moreover, we have used an ensemble of models to address the issue that the cross attention matrices usually differ across different random initializations. However, as suggested in \cite{brunner2019identifiability}, the identifiability of the cross attention needs to be examined more closely. Additionally, several important biological features, such as synaptic plasticity and neuromodulation, are not yet included in the current simulations, and should be implemented in future experiments. Finally, caution must be taken when interpreting causal relationships inferred by Granger causality, as predictability does not always imply causality, especially when not all relevant variables are observed. Despite these limitations, our work highlights the potential of the transformer model in causal representation learning in neuroscience, inviting further exploration of its applications in unraveling the complexities of neuronal networks and deepening our understanding of brain function.

\section*{Acknowledgements}
We thank Praveen Venkatesh and Che Wang for the helpful discussions. 

We acknowledge the partial support of NSF Grant NSF NCS-FO \#2024364

This research was supported in part by an appointment with the National Science Foundation (NSF) Mathematical Sciences Graduate Internship (MSGI) Program. This program is administered by the Oak Ridge Institute for Science and Education (ORISE) through an interagency agreement between the U.S. Department of Energy (DOE) and NSF. ORISE is managed for DOE by ORAU. All opinions expressed in this paper are the author's and do not necessarily reflect the policies and views of NSF, ORAU/ORISE, or DOE. 

This manuscript has been authored by UT-Battelle, LLC, under contract DE-AC05-00OR22725 with the US Department of Energy (DOE). The US government retains and the publisher, by accepting the article for publication, acknowledges that the US government retains a nonexclusive, paid-up, irrevocable, worldwide license to publish or reproduce the published form of this manuscript, or allow others to do so, for US government purposes. DOE will provide public access to these results of federally sponsored research in accordance with the DOE Public Access Plan (https://www.energy.gov/doe-public-access-plan).

\newpage
\bibliography{ref}

\newpage
\appendix

\section{Technical Details}
\subsection{Synthetic data generation}

We generated the synthetic datasets using the Izhikevich model~\cite{izhikevich2003simple}
\begin{align*}
    v' &= 0.04v^2 + 4.1v + 108 - u + I \\
    u' &= a(bv-u)
\end{align*}
with the auxiliary after-spike resetting mechanism
\begin{equation*}
    \text{if } v \geq 30mV, \text{then } \begin{cases}
        v \leftarrow c \\
        u \leftarrow u + d
    \end{cases}
\end{equation*}
Values for the parameters $a, b, c, d$ for both excitatory and inhibitory neurons are listed in Table \ref{izh}. 

\begin{table}[h!]
  \centering
\begin{threeparttable}
  \caption{Parameters of Izhikevich model}
  \label{izh}
  \begin{tabular}{|l|l|l|l|l|}
    \toprule
    \diagbox{Neuron type}{Parameter} & $a$ & $b$ & $c$ & $d$ \\
    \midrule
    Excitatory & $0.02$ & $-0.1$ & $-65+15\theta^2$ & $8-6\theta^2$ \\ \hline
    Inhibitory & $0.02 + 0.08\theta$ & $-0.1$ & $-65$ & $2$ \\
    \bottomrule
  \end{tabular}
  \begin{tablenotes}
      \item where $\theta \sim \textit{Uniform}(0, 1)$.
    \end{tablenotes}
\end{threeparttable}
\end{table}

\subsection{Additional Details on the Causalformer Architecture}
\begin{figure}[h!]
  \centering
  \includegraphics[width=0.8\textwidth]{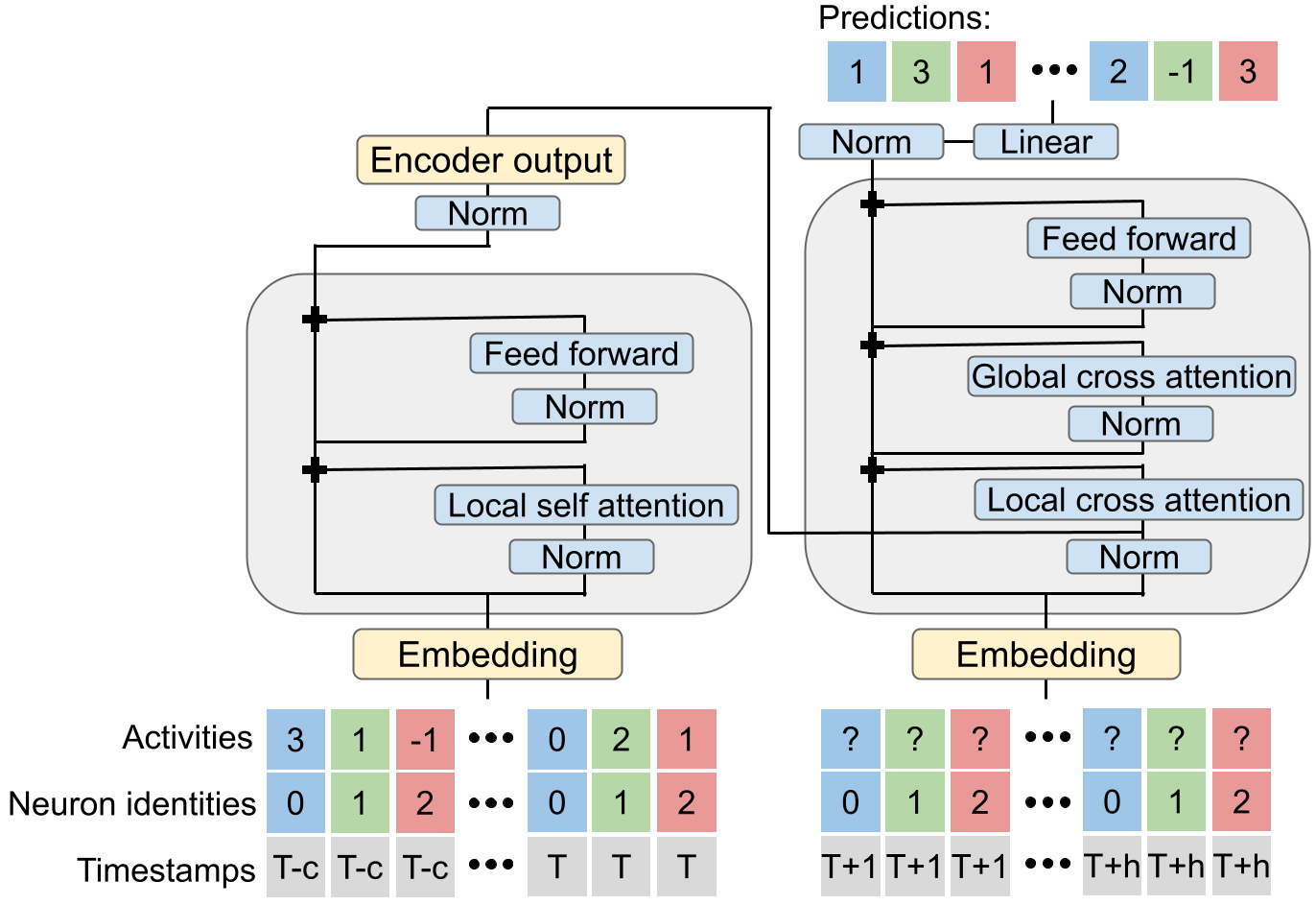}
  \caption{Causalformer architecture. In encoder local self attention, each neuron can only attend to the embedding of its own history. Similarly, in decoder local cross attention, the target embedding of each neuron can only attend to the encoder representation of its own history. In decoder global cross attention, the target embedding of each neuron can attend to the encoded history representations of all neurons, including itself. The distinction between local and global attentions was first introduced in \cite{grigsby2021long} as an architectural bias. Causal relationships between neurons are inferred from the decoder's global cross attention module. }
  \label{model}
\end{figure}

Figure \ref{model} shows the details of the encoder-decoder architecture of Causalformer. The architecture and the illustration are based on and modified from \cite{grigsby2021long}. 

\subsection{Details of the attention mechanism}\label{sec:attn}
Following \cite{vaswani2017attention}, the output of the attention module in Causalformer is computed based on a set of keys, queries, and values. Let $K \in \mathbb{R}^{L_K \times d_K}$ be a matrix of keys, $Q \in \mathbb{R}^{L_Q \times d_K}$ be a matrix of queries, and $V \in \mathbb{R}^{L_K \times d_V}$ be a matrix of values, where each row is an individual key/query/value. In our implementation we take $d_K = d_V = d$. The output of the attention module is 
\begin{equation*}
    \mathcal{A}(Q, K, V) = \text{softmax}(\frac{QK^T}{\sqrt{d}})V
\end{equation*}
Specifically, consider using Causalformer to predict an $h$-step target sequence $(x_{t+1}, \dots, x_{t+h})$ from a $c+1$-step history sequence $(x_{t-c}, \dots, x_t)$. Encoder local self attention is computed for each neuron separately, where $Q, K, V$ are linear projections of its own history sequence embedding. Since each timestep is embedded as separate token, each neuron has a history sequence of $c+1$ tokens, so $L_K = L_Q = c+1$. Decoder local cross attention is also computed for each neuron separately, where $Q$ is a linear projection of its target sequence embedding, and $K, V$ are linear projections of its history sequence embedding. Since there are $h$ timesteps to predict, each neuron has a target sequence of $h$ tokens, so $L_Q = h$, $L_K = c+1$. Decoder global cross attention is computed for all neurons simultaneously, where $Q$ is a linear projection of all neurons' target sequence embedding, $L_Q = Nh$, and $K, V$ are linear projections of all neurons' history sequence embedding, $L_K = N(c+1)$. 

\subsection{Evaluation with AUROC}\label{sec:auroc}
For the decoder cross attention matrix $A$ of Causalformer, $A_{ij}$ reflects how important neuron $j$'s history is for the prediction of neuron $i$'s future, or in other words, how likely that there exists a causal relationship between neuron $j$ and neuron $i$. Since the ground-truth connectivity matrix is binary, one way to assess the degree to which $A$ reflects the ground-truth causality is to specify a minimum importance needed for establishing a causal relationship. That is, if $A_{ij}$ is greater than a certain threshold, then we say neuron $j$ Granger-causes neuron $i$. However, as crucial as the threshold is, choosing a fixed number as the threshold can introduce additional arbitrariness. As an alternative to a fixed threshold, the Receiver Operating Characteristic curve (ROC curve) plots the true positive rate (i.e., a causal relationship is correctly identified) versus the false positive rate (i.e., a causal relationship is established where it actually doesn't exist) at a range of different thresholds. In a Cartesian coordinate system, all ROC curves start at $(0, 0)$, and end at $(1, 1)$. Visually, the better $A$ corresponds to the ground-truth connectivity, the more rapidly ROC curve rises to 1, and the closer the area under the ROC curve is to 1. Therefore, we used AUROC, area under the ROC curve, to quantify the degree to which $A$ reflects the ground-truth connectivity. 

In MVGC, pairwise causal relationships are also inferred from an $N \times N$ matrix, which we denote by $F$, where $F_{ij} > 0$ reflects the degree to which neuron $j$ helps predicting the future of neuron $i$ (see \cite{barnett2014mvgc} for details of how $F_{ij}$ is computed). In short, larger $F_{ij}$ gives stronger evidence for neuron $j$ Granger-causing neuron $i$. In the MVGC routine, the next step is to determine the statistical significance of the pairwise causal relationships from $F_{ij}$, and only relationships with sufficient significance (i.e., when $F_{ij}$ is large enough) are reported. However, since all potential relationships, both weak and strong, are reported in the cross attention matrix, to ensure a fair comparison, instead of checking only statistical significant relationships from MVGC, we computed AUROC using $F$ directly, considering all $F_{ij}$ regardless of its magnitude. The applicability of AUROC to the evaluation of both $A$ and $F$ is based on two key factors: (a) in the ground-truth connectivity matrix, 1 represents existent connections and 0 represents no connection, (b) both larger $A_{ij}$ and $F_{ij}$ correspond to more confidence for the existence of connections.

\clearpage

\section{Additional Results}

\begin{figure}[htb]
  \centering
  \includegraphics[width=0.99\textwidth]{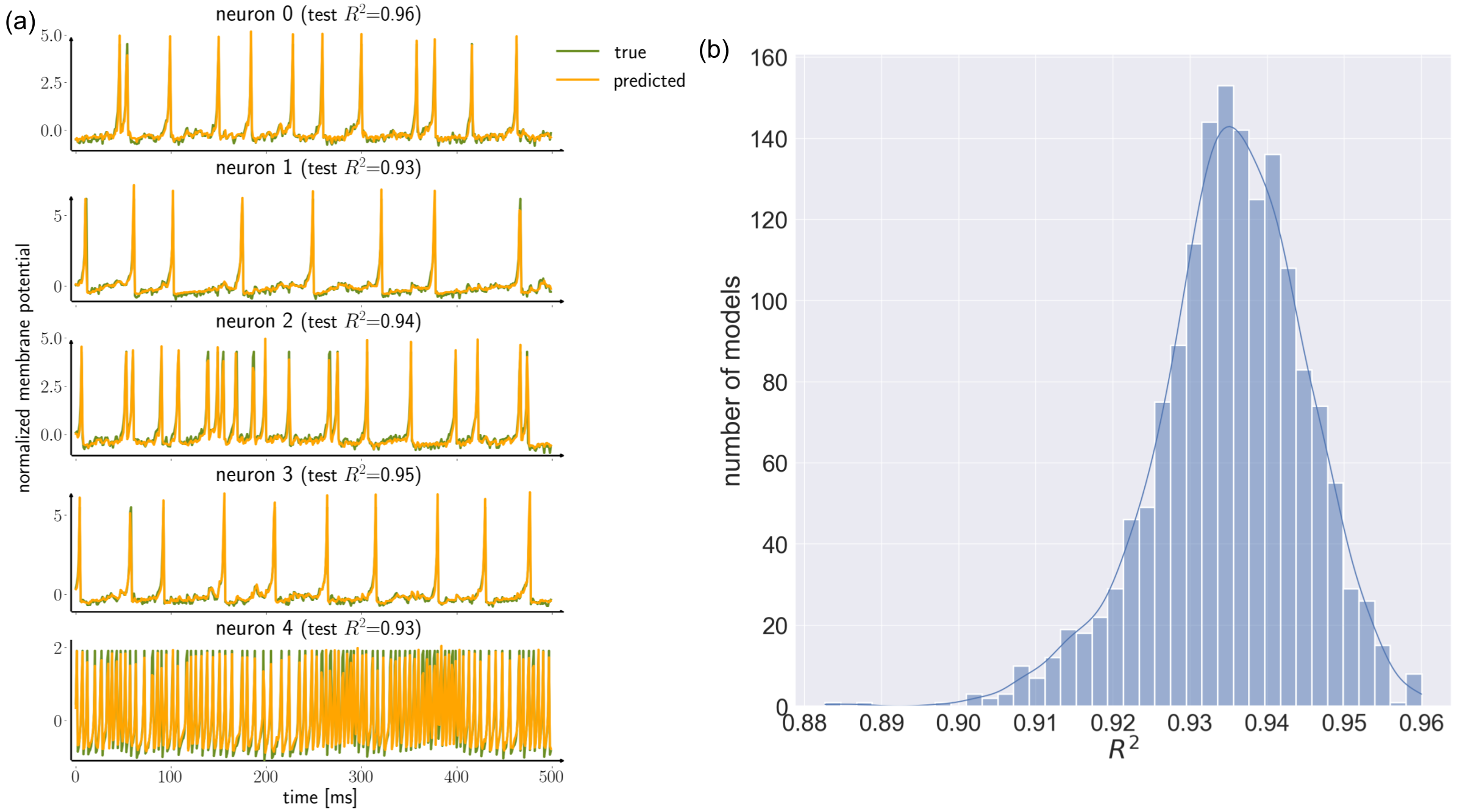}
  \caption{Causalformer prediction quality. \textbf{(a)} Example true and predicted normalized membrane potential traces. Since we make one-step prediction in each test sample, predicted traces are made by concatenating predictions across test samples. Test $R^2$ scores are computed between the true and predicted traces on the test set. \textbf{(b)} Distribution of test $R^2$ scores (averaged across all neurons) achieved by all Causalformer models trained on the random networks. Note that test $R^2$ scores are concentrated around 0.935, similar to those in (a), which means that the vast majority of the models are able to capture the dynamics reasonably well, based on the visualization in (a)}
  \label{pred}
\end{figure}

Figure \ref{pred} provides a visualization of Causalformer predition. Qualitatively, we confirm that Causalformer is able to capture the dynamics reasonably well.

\section{Hyperparameters}
Table \ref{hyper} lists the hyperparameters used for training Causalformer. The same set of hyperparameters was used for all random network datasets. For each dataset we trained 10 different Causalformer models with 10 random seeds. 

\begin{table}[h!]
  \caption{Hyperparameters of Causalformer}
  \label{hyper}
  \centering
  \begin{tabular}{ll}
    \toprule
    Hyperparameter              & Value   \\
    \midrule
    Embedding dimension                  & 100  \\
    Feed-forward network dimension   & 400 \\
    Time embedding dimension         & 1 \\
    Query, key, value dimension      & 8 \\
    Number of attention heads        & 10  \\
    Number of decoder layer          & 1 \\
    Number of encoder layer          & 1 \\
    Embedding dropout probability    & 0.1 \\
    Feed-forward dropout probability & 0.1 \\
    Batch size                       & 16 \\
    Initial learning rate            & 0.0005 \\
    Learning rate decay factor       & 0.5 \\
    AdamW weight decay coefficient   & 0.001 \\
    Maximum number of epochs         & 200 \\
    Number of no-improvement epochs to trigger early stopping & 10 \\
    \bottomrule
  \end{tabular}
\end{table}


\end{document}